\documentclass[10pt,twocolumn,letterpaper]{article}

\usepackage{cvpr}

\usepackage{graphicx}
\usepackage{amsmath}
\usepackage{amssymb}
\usepackage{booktabs}

\usepackage{lipsum}
\usepackage{multirow}
\usepackage[table]{xcolor}
\usepackage{pifont}
\usepackage{subcaption}
\newcommand{\cmark}{\ding{51}}

\usepackage{color}

\usepackage[pagebackref,breaklinks,colorlinks]{hyperref}

\makeatletter
\def\thanks#1{\protected@xdef\@thanks{\@thanks
        \protect\footnotetext{#1}}}
\makeatother

\usepackage[capitalize, noabbrev]{cleveref}

\def\cvprPaperID{2476} 
\def\confName{CVPR}
\def\confYear{2023}

\begin{document}

\title{MixTeacher: Mining Promising Labels with Mixed Scale Teacher\\ for Semi-Supervised Object Detection}

\author{Liang Liu$^{1}$,~~~Boshen Zhang$^{1}$,~~~Jiangning Zhang$^{1}$, ~~~Wuhao Zhang$^{1}$,~~~Zhenye Gan$^{1}$\\Guanzhong Tian$^{3}$, ~~~Wenbing Zhu$^{4}$,~~~Yabiao Wang$^{1\dag}$\thanks{$^\dag$ Corresponding Authors.},~~~Chengjie Wang$^{1, 2\dag}$, 
\\ 
$^1$Youtu Lab, Tencent~~~$^2$Shanghai Jiao Tong University\\
$^3$Ningbo Research Institute, Zhejiang University, ~~~$^4$Rongcheer Co., Ltd\\
\tt\small $\{$leoneliu, boshenzhang, vtzhang, wuhaozhang, wingzygan$\}$@tencent.com;\\
\tt\small gztian@zju.edu.cn; louis.zhu@rongcheer.com; $\{$caseywang, jasoncjwang$\}$@tencent.com
}
\maketitle

\begin{abstract}
    Scale variation across object instances remains a key challenge in object detection task. Despite the remarkable progress made by modern detection models, this challenge is particularly evident in the semi-supervised case. While existing semi-supervised object detection methods rely on strict conditions to filter high-quality pseudo labels from network predictions, we observe that objects with extreme scale tend to have low confidence, resulting in a lack of positive supervision for these objects. In this paper, we propose a novel framework that addresses the scale variation problem by introducing a mixed scale teacher to improve pseudo label generation and scale-invariant learning. Additionally, we propose mining pseudo labels using score promotion of predictions across scales, which benefits from better predictions from mixed scale features. Our extensive experiments on MS COCO and PASCAL VOC benchmarks under various semi-supervised settings demonstrate that our method achieves new state-of-the-art performance. The code and models are available at \url{https://github.com/lliuz/MixTeacher}.

\end{abstract}

\vspace{-1em}
\section{Introduction}
\begin{figure}[t]
	\centering
\vspace{1em}
	\includegraphics[width=0.98\columnwidth]{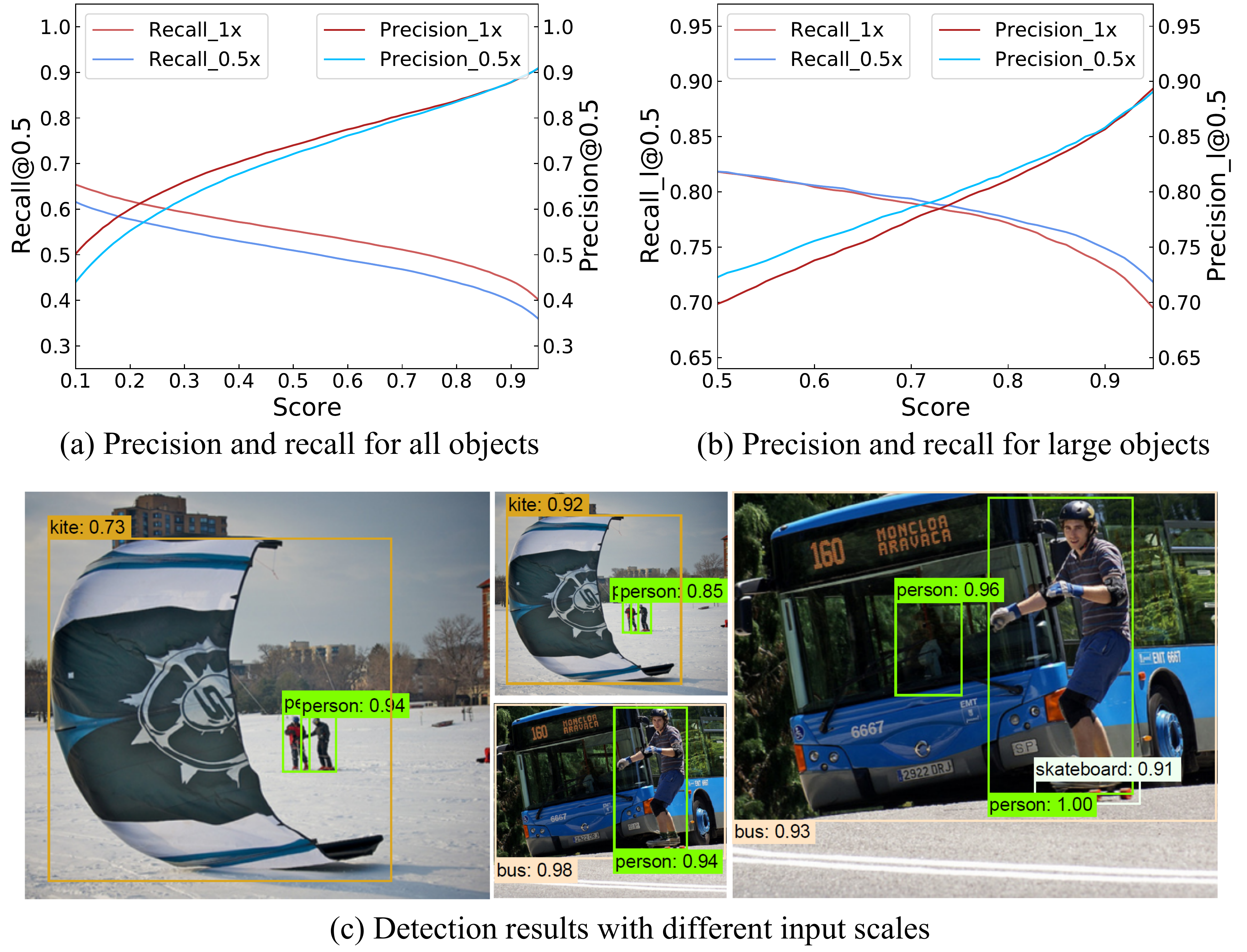}
    \vspace{-.5em}
	\caption{Detection results with input of regular 1$\times$ scale and 0.5$\times$ down-sampled scale images.  We plot the precision and recall under different score thresholds for (a) all objects and (b) large objects in COCO \texttt{val2017} with the same model but different input scales. Two examples of unlabeled images are given in (c). Large scale inputs have clear advantages in overall metrics, but down-sampled images are more suitable for large objects.}
	\vspace{-.5em}
	\label{fig:f1}
\end{figure}

The remarkable performance of deep learning on various tasks can largely be attributed to large-scale datasets with accurate annotations. However, collecting a large amount of high-quality annotations is infeasible as it is labor-intensive and time-consuming, especially for tasks with complicated annotations such as object detection~\cite{coco, objects365} and segmentation~\cite{pascalvoc, cityscapes}. To reduce reliance on manual labeling, semi-supervised learning~(SSL) has gained much attention. SSL aims to train models on a small amount of labeled images and a large amount of easily accessible unlabeled data. Following extensive pioneering studies on semi-supervised image classification~\cite{Fixmatch,mixmatch,temporalensembling}, several methods on semi-supervised object detection have emerged.

Most early studies on semi-supervised object detection~\cite{STAC,unbiasedTeacher,CSD} can be considered as a direct extension of SSL methods designed for image classification, using a teacher-student training paradigm~\cite{MeanTeacher,Fixmatch,mixmatch}. In these methods, a teacher model generates pseudo bounding boxes and corresponding class predictions on unlabeled images, and the pseudo labels are used to train a student model. Despite the performance improvement from using a large amount of unlabeled data, these methods overlooked the characteristics of object detection to some extent, resulting in a huge gap from the fully supervised counterpart.

\begin{figure*}[t]
	\centering
	\includegraphics[width=0.99\textwidth]{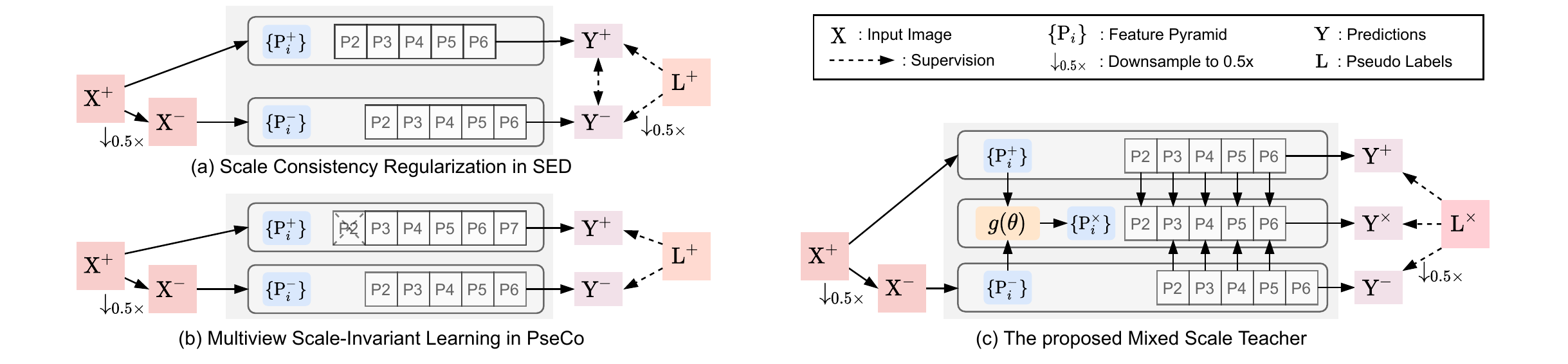}
    \vspace{0em}
	\caption{Comparison of multi-scale learning in semi-supervised object detection methods. Previous methods~\cite{sed, pseco} only focus on encouraging consistent predictions for the input image with different scales. The proposed MixTeacher explicitly introduces a mixed scale feature pyramid to adaptive fuse features from appropriate scales, which is capable to detect objects with varying sizes. The mixed scale features generate more accurate pseudo labels and help to mine promising labels, as a plug-in which can be discarded after training.}
	\vspace{-0.8em}
	\label{fig:method_cmp}
\end{figure*}

Compared to image classification, object instances in detection tasks can vary in a wider range of scales. To address this challenge of detecting and localizing multiple objects across scales and locations, numerous works in object detection have been proposed, such as FPN~\cite{fpn}, Trident~\cite{trident}, and SNIP~\cite{snip}. However, the large scale variation brings new challenges in the semi-supervised context. In order to guarantee high precision, most existing semi-supervised object detection methods adopt strict conditions (\eg score $>$ 0.9) to filter out highly confident pseudo labels. Although this ensures the quality of pseudo labels, many objects with low confidence are wrongly assigned as background, especially for those with extreme scales. As shown in~\cref{fig:f1} (c), inappropriate scales will lead to false negatives, which can mislead the network in semi-supervised learning.
We further observe the influence of the test scale of the images. Consistent with common sense, large-scale inputs have clear advantages in overall metrics, as shown in~\cref{fig:f1}~(a). However, down-sampled images show a superiority for large objects, as shown in~\cref{fig:f1} (b). This provides a new view to handle the scale variation issue.

It is worth mentioning that recent works have paid attention to the scale variation issue in semi-supervised object detection. As shown in~\cref{fig:method_cmp} (a) and (b), previous methods have introduced an additional down-sampled view to encourage the model to make scale-invariant predictions. Specifically, SED~\cite{sed} proposes to distill predictions of class probability from the regular scale to the down-sampled scale and constrain consistent predictions of localization for all proposals in two scales. PseCo~\cite{pseco} adopts the same pseudo labels generated from the regular scale for both scales. However, these methods mainly focus on the consistency of predictions across scales, which indirectly improves the models with regularization. Moreover, they highly rely on the pseudo labels generated from the regular scale in the teacher network. The false negatives caused by inappropriate scales still remain in these methods.

Based on the above methods, which are equipped with an additional down-sampled view of unlabeled images, we propose to explicitly improve the quality of pseudo labels to handle the scale variation of objects. As shown in~\cref{fig:method_cmp}~(c), we introduce a mixed-scale feature pyramid, which is built from the large-scale feature pyramid in the regular view and the small-scale feature pyramid in the down-sampled view. The mixed-scale feature pyramid is supposed to be capable of adaptively fusing features across scales, thus making better predictions in the teacher network. Furthermore, to avoid object instances missing in the pseudo labels due to low confidence scores, we propose to leverage the improvement of score as an indicator for mining pseudo labels from low confidence predictions. In summary, the main contributions are as follows:
\vspace{-.5em}
\begin{itemize}
    \item We propose a semi-supervised object detection framework MixTeacher, in which high-quality pseudo labels are generated from a mixed scale feature pyramid. 
    \vspace{-.5em}
    \item We propose a method for pseudo labels mining, which leverages the improvement of predictions as the indicator to mining the promising pseudo labels.
    \vspace{-.5em}
    \item Our method achieves state-of-the-art performance on MS COCO and Pascal VOC benchmarks under various semi-supervised settings. 
\end{itemize}

\section{Related works}
\noindent \textbf{Semi-supervised Learning} aims to train a model using a small amount of labeled data and a large amount of unlabeled data. Early studies mainly focused on image classification task~\cite{temporalensembling, Fixmatch} and have gradually been generalized to various tasks~\cite{ouali2020semi,STAC,zhao2020sess}. Pseudo labeling~\cite{pseudo-label1} is one of the most popular paradigms in semi-supervised image classification, where labels of unlabeled data are generated by a pre-trained teacher network. Under this paradigm, Mean Teacher~\cite{MeanTeacher} proposes to maintain the teacher model as an exponential moving average~(EMA) of the student model, thus generating pseudo labels end-to-end. Meanwhile, some works encourage models to make consistent predictions with perturbations~\cite{miyato2018virtual,bachman2014learning,sajjadi2016regularization}. Typical approaches such as MixMatch~\cite{mixmatch} and UDA~\cite{UDA} enforce consistent predictions across image views with different augmentations. Our work follows the pseudo labeling paradigm and consistency regularization, but focuses on object detection, where some task-specific challenges are relatively under-explored.

\vspace{.5em}
\noindent \textbf{Semi-supervised Object Detection (SSOD)} methods are mainly derived from semi-supervised image classification. As early attempts, CSD~\cite{CSD} encourages consistent predictions for horizontally flipped image pairs, whereas STAC~\cite{STAC} transfers the weak and strong augmentations from FixMatch~\cite{Fixmatch} to SSOD. After that, ~\cite{instantTeaching,humbleTeacher,ISMT,softTeacher} simplify the trivial multi-stage training with the idea of EMA from Mean Teacher~\cite{MeanTeacher}, realizing the end-to-end training. Considering the characteristic of object detection, some task-specific improvements are proposed. Soft Teacher~\cite{softTeacher} adopts separate and strict conditions to filter out high-quality pseudo labels for classification and regression and reduces the classification weights of negative proposals to suppress the influence of missing objects in pseudo labels. Unbiased Teacher~\cite{unbiasedTeacher} replaces the cross-entropy loss with Focal loss~\cite{focal} to alleviate the numerous negative pseudo labels problem. On the basis of these work, our work focus on another challenge in SSOD, \ie the large scale variation problem. 

\vspace{0em}
\noindent \textbf{Scale Variation Challenge} exists in most vision tasks. Since the scale of object instances in the detection task could vary in a wide range, numerous methods were proposed to detect objects across scales~\cite{fpn,trident,snip}. 
However, scale variation brings new challenges for semi-supervised object detection. For instance, objects with extreme scale tend to have low confidence, which makes them missed in the pseudo labels for most SSOD methods with a strict filter condition~\cite{STAC, softTeacher}. Recent attempts have noticed the scale problem.~\cite{sed} proposed to distill the predictions between a regular view and a down-sampled view to enhance the robustness of the model for scale variation.~\cite{pseco} also adopt a down-sampled view but shifts the layer of feature pyramid to reuse the same scale pseudo boxes as a regular view. These methods could be regarded as adding consistent regularization on varying scales. In contrast, our work proposes a mixed scale feature pyramid which can adaptive select scale for generating and mining promising pseudo labels.

\begin{figure*}[t]
	\centering
	\includegraphics[width=0.95\textwidth]{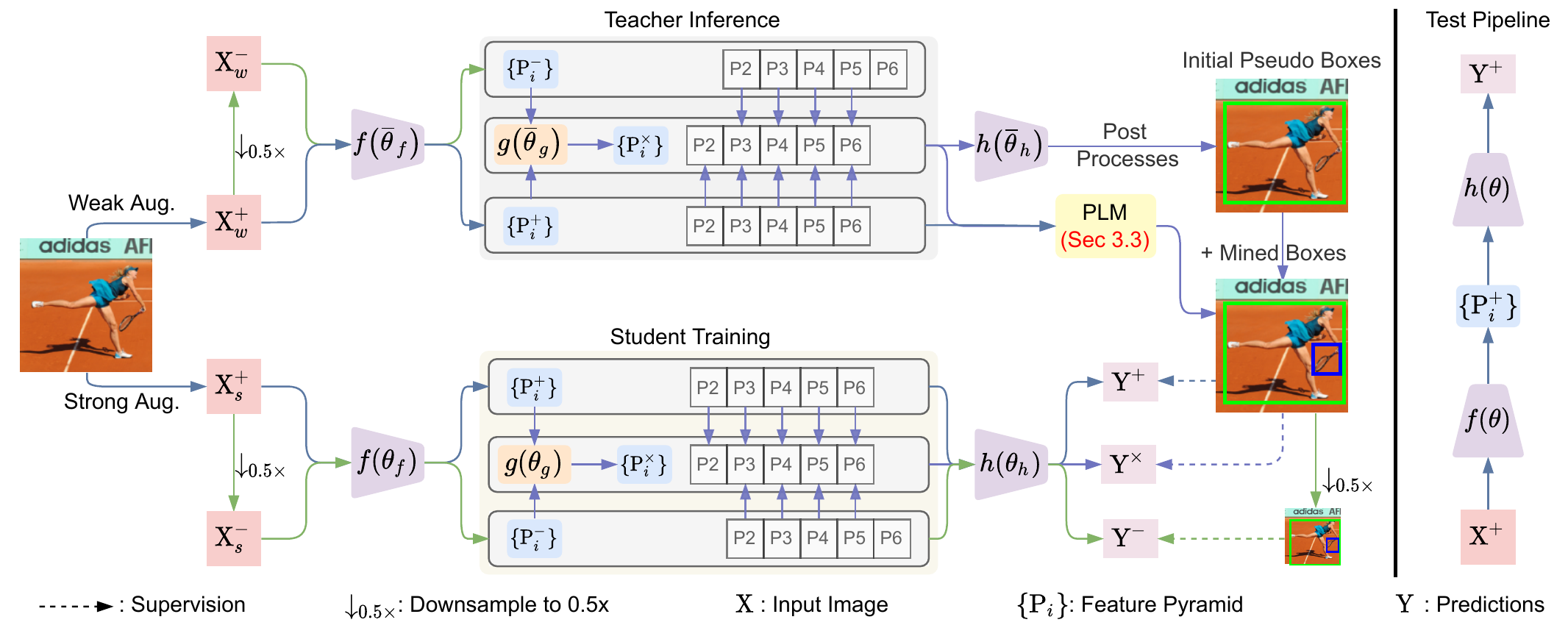}
    \vspace{-.5em}
	\caption{During training, the model first constructs two feature pyramids for a regular scale and a down-sampled scale with a feature extraction module $f(\theta_f)$, respectively. Next, an additional mixed scale feature pyramid is built by a feature fusion module $g(\theta_g)$. The student model trains on three scales with a shared detection head $h(\theta_h)$ taking the pseudo boxes generated from the mixed-scale of teacher model as supervisions. In addition, promising labels with low confidence scores are mined with a PLM strategy. The weights $\bar\theta$ in teacher are updated by EMA of the weights $\theta$ in student. 
 In testing, the model with original architecture and 
 regular input scale is used.
 }
	\vspace{-.5em}
	\label{fig:pipeline}
\end{figure*}
\section{Method}

In the problem of semi-supervised object detection, a model is trained with a labeled set $\mathcal{D}^l=\{(\text{X}^s_i, \text{Y}^s_i)|_{i=1}^{N^l}\}$ and an unlabeled image set $\mathcal{D}^u=\left\{\text{X}^u_i|_{i=1}^{N^u}\right\}$, where $N^l$ and $N^u$ are the numbers of labeled and unlabeled data.
For each labeled image $\text{X}^s_i$, the annotation $\text{Y}^s_i$ is composed of a set of boxes and corresponding category labels for the instances that appeared in the image. Built upon the pseudo labeling framework, our method follows a score filtering mechanism~\cite{softTeacher} to generate pseudo labels $\text{Y}^u$ for unlabeled image $\text{X}^u$. An overview of our method is shown in~\cref{fig:pipeline}. We propose a mixed scale feature pyramid for pseudo labels generation and scale consistent learning, and based on it, a promising label mining strategy is introduced. Note that although the proposed method is independent of detection models, we adopt Faster RCNN~\cite{faster} with FPN~\cite{fpn} as the default model, following most of the previous work.

\subsection{Basic Pseudo-Labeling Framework}
\label{sec:basic_framwork}
Following the common practice in previous work~\cite{softTeacher, unbiasedTeacher, pseco}, we adopt the pseudo-labeling under the teacher-student paradigm as our basic training framework. Specifically, the training images are sampled from both labeled and unlabeled datasets, and the overall objective is made up of these two parts to update a student model. Due to the lack of ground truth on unlabeled images, a teacher model provides pseudo labels for students, whose weights are updated by the exponential moving average of the student model.

In every training iteration, the training objective on labeled data follows a regular manner with fully supervised by the ground truth labels. On the unlabeled data, the teacher model first generates pseudo labels on a weakly augmented view of the image, which provides supervision signals for a strongly augmented view of the image for the student model. Subsequently, the student model is updated with the objective from labeled data and a strongly augmented view of the image with pseudo labels. The overall training objective can be formulated as:
\begin{equation}
\mathcal{L}=\mathcal{L}_{s}+\alpha \mathcal{L}_{u}    
\end{equation}
where $\mathcal{L}_{s}$ and $\mathcal{L}_{u}$ denote supervised loss of labeled images and unsupervised loss of unlabeled images respectively, $\alpha$ controls contribution of unsupervised loss.

\subsection{Mixed Scale Teacher}

Existing works have proved that incorporating an extra down-sampled view of the unlabeled image, and regularizing the network with consistency constraints on either feature level or label level can significantly improve the performance of semi-supervised object detection~\cite{sed, pseco}. Based on this observation, we also leverage the down-sampled view, but resort to building a more informative feature representation, which is more suitable for pseudo labels. 

Given an image, most of detectors first extract multi-scale features $P_i$ with decreasing spatial sizes which constitute a feature pyramid $\mathbb{P}$. In the case of FPN~\cite{fpn}, the spatial sizes of adjacent levels in the feature pyramid always differ by $2\times$, which results in $P_2-P_6$ layers\footnote{In the original implementation of Faster RCNN, P2-P6 is used, while in some detector such as FCOS and RetinaNet, P3-P7 is used.} with spatial sizes from $1/{2^2}$ to $1/2^6$ \wrt the size of the input image. 

In this work, we first extract two feature pyramids from the regular view and the down-sampled view of the input image, respectively, which denote $\mathbb{P}^+=\{P_2^+,..., P_6^+\}$ and $\mathbb{P}^-=\{P_2^-,..., P_6^-\}$, respectively. Then, we build a mixed scale feature pyramid $\mathbb{P}^{\times}=\{P_2^\times,..., P_6^\times\}$ from the aforementioned two views to adaptive fuse appropriate features for input images containing objects of different scales. 

Notice that with a 0.5$\times$ down-sample ratio, the network produces a small scale feature pyramid. Specifically, the feature in $P^+_i$ shares the same spatial size as $P^-_{i - 1}$. The spatial aligned feature maps of adjacent levels in two scales greatly simplify the design of feature fusion module. To avoid introducing a complex feature space, we propose a linear combination of features from the regular and down-sampled scales for the mixed scale feature pyramid, where the weight is formulated as follows:
\begin{equation}
    \gamma = \sigma\left(g\left(P_i^+,P_{i-1}^-|\theta_g\right)\right),
\end{equation}
where $\sigma(.)$ denotes the sigmoid activation with temperature $T$, and $g(.| \theta_g)$ is inspired by the SE block~\cite{SENet} which involves global average pooling followed by MLPs for channel concatenation of $P_i^+$ and $P_{i-1}^-$. Thus, the feature $P^\times_{i}$ in the mixed scale feature pyramid is a weighted sum of large scale and small scale features, which can be derived as:
\begin{equation}
    P^\times_{i} = \gamma P^+_{i} + (1 - \gamma) P^-_{i - 1}.
\end{equation}
Note that the first level in the mixed-scale pyramid is a direct copy from regular view, for which the corresponding level in the down-sampled view does not exist.

As the mixed scale feature pyramid contains more information, we use it to generate pseudo labels for unlabeled images in the teacher network. Consequently, the training objective for unlabeled data becomes:
\begin{equation}
\begin{split}
\mathcal{L}_{u} =& \mathcal{L}_{det}\left(h(\mathbb{P}^{\times}), \hat{y}^{\times}\right) + \mathcal{L}_{det}\left(h(\mathbb{P}^+), \hat{y}^\times\right) \\+& \mathcal{L}_{det}\left(h(\mathbb{P}^-), \hat{y}^\times\right),
\end{split}
\end{equation} 
in which the overall loss is the sum of losses from all scales. Similarly, the objective of the labeled part are also extended to multiple scales. For the specific form of $\mathcal{L}_{det}$ at each scale, we followed the baseline method~\cite{softTeacher}.

It is worth mentioning that after extracting the regular and down-sampled scale features, the extra computation and parameters for building the mixed scale feature pyramid are negligible. Specifically, we implement the feature fusion module using two linear layers and a ReLU activation, which is a drop in the bucket compared to the millions of parameters in modern detection models.

Once the model is trained completely, all components not belong to the original detector can be discarded, including the steps used to extract small/mixed scale features and the feature fusion module. It satisfies the method to maintain the original detector architecture and input scale for inference, ensuring fairness for comparison.

\subsection{Promising Labels Mining}

In most of the previous methods, a high score threshold $\tau_h$ is applied to filter out the high-quality pseudo labels from the predictions of teacher network. However, it brings false negatives due to the inappropriate scales which will mislead the student network in semi-supervised learning.

To address this problem, we propose a strategy named Promising Labels Mining (PLM), which takes the predictions from the mixed scale feature pyramid as the target, and measure the improvement from the regular scale. The improvement of confidence score is adopted as the indicator to mine pseudo labels from predictions with score $\tau_l < s \leq \tau_h $, which we called low-confidence candidates.

As a common practice in label assignment methods for fully supervised object detection~\cite{tood, ota, MutualSup, fcos, dwla}, we first construct a bag of proposals from RPN for each low-confidence candidate, which will be used to measure the quality of each candidate. To reduce training costs, proposals that do not belong to any bag of low-confidence candidates or are not assigned to any high-confidence pseudo labels are not involved in the process of mining promising labels.

Notably, in order to avoid the influence introduced by strong augmentation, the proposals and feature pyramids from the teacher network are used to measure the score improvement. Besides, only the regular view is adopted as the source view, and the mixed scale is adopted as the target view to mine promising pseudo labels from candidates. The proposals generated from the source view are shared with the target view to align the input of the two views.

For a candidate with the highest score on class $c$, we define the indicator $\Delta q$ as the promotion of the mean score from the source to the target view, formulated as:
\begin{equation}
\Delta q = \frac{1}{K}\left(\sum_{i=1}^Kh_c(b_i, \mathbb{P}^\times | \bar\theta) - \sum_{i=1}^Kh_c(b_i, \mathbb{P}^+ | \bar\theta)\right)    
\end{equation}
where $K$ is the number of proposals in the bag for this candidate. $h_c(.|\bar\theta)$ denotes the classification head in the teacher network, which takes the feature pyramid $\mathbb{P}$ and the RoI of each proposal $b$ as input, and predicts the confidence score of class $c$ for each proposal in the bag. 

A low-confidence candidate with a higher $\Delta q$ indicates the corresponding object instance prediction could reach greater promotion in the mixed scale feature pyramid. Therefore, it may be hard to detect in the source view but relatively easy in the target view with more information, suggesting that it is more likely to be a hard instance. 

Candidates with $\Delta q$ higher than the promotion threshold $\delta$ are regarded as pseudo labels. These mined pseudo labels, along with the existing high-confidence pseudo labels, are used in the classification loss for all three scales.

\textbf{Discussion:} Previous works~\cite{pseco, tood, ota} mainly utilize the idea of the bag of proposals for label assignment to reduce the proposals wrongly assigned to false positives or false negatives, which cannot address the missing pseudo labels with low confidence. In this work, we focus on mining low-confidence pseudo labels, and propose to measure their quality with two views. Therefore, our method is orthogonal to those label assignment methods.
Besides, we only mined pseudo labels for the classification loss, while the criterion of filtering pseudo labels for the regression loss follows a widely used baseline~\cite{softTeacher}, in which an uncertainty threshold is adopted for each box prediction. Other advanced strategies for filtering localization pseudo labels~\cite{pseco, VCL} can also be integrated with our method, which might lead to better results but are irrelevant to our mixed scale framework, thus we do not discuss them in this paper.

\section{Experimental Results}
\subsection{Dataset and Evaluation Protocol}

We present experimental results on MS COCO~\cite{coco} and Pascal VOC~\cite{pascalvoc} benchmarks. Specifically, there are three common settings following previous works~\cite{softTeacher, sed}. The results are evaluated on \texttt{val2017} for the settings on COCO, and VOC17 \texttt{test} set for the setting on VOC, respectively.
The specific data settings are as follows:

\noindent \textbf{COCO Partially Labeled.} It contains 4 splits with various labeled ratios, in which 1\%, 2\%, 5\%, and 10\% images from \texttt{train2017} are randomly sampled as labeled data, and the rest images are used as the unlabeled part for each split, respectively. Following the common practice, 5 different folds are used to evaluate each split, and the mean and standard deviation of results are reported.

\noindent \textbf{COCO Additional.} All 118k images in \texttt{train2017} are adopted as labeled data, and all of the additional 123k images in \texttt{unlabeled2017} are used as unlabeled data.

\noindent \textbf{VOC Additional.} VOC07 \texttt{trainval} is used as labeled data and VOC12 \texttt{trainval} is used as unlabeled data.

\noindent \textbf{VOC Mixture.} VOC07 \texttt{trainval} is used as labeled data and images from COCO containing 20 classes in VOC along with VOC12 \texttt{trainval} are taken as unlabeled data.

The evaluation metrics of this paper including AP at different IoU thresholds (\eg $\text{AP}_{50:95}$ denoted as mAP, $\text{AP}_{50}$, $\text{AP}_{75}$) and different box scales (\eg $\text{AP}_{s}$, $\text{AP}_{m}$, $\text{AP}_{l}$). All metrics are calculated via the COCO evaluation kit.

\begin{table*}[t]
    \centering
    \resizebox{0.76\linewidth}{!}{
    \renewcommand{\arraystretch}{1.05}
    \begin{tabular}{p{4.cm}|c|c|c|c|c}
    \toprule
      & \multicolumn{4}{c|}{COCO Partially Labeled} &  COCO Additional \\ \cmidrule{2-5} 
     & 1\% & 2\% & 5\% & 10\% & 100 \% \\ \midrule
    Supervised Baseline & 12.15$\pm$0.27 & 16.65$\pm$0.18 & 21.45$\pm$0.16 & 27.10$\pm$0.07 & 40.9 \\ \midrule
    STAC~\cite{STAC} &  13.97$\pm$0.35 & 18.25$\pm$0.25 & 24.38$\pm$0.12 & 28.64$\pm$0.21 & 39.5 $\xrightarrow{-0.3}$ 39.2 \\
    SED~\cite{sed}  & - & - & 29.01 & 34.02 & 40.2 $\xrightarrow{+3.2}$ 43.4 \\

    Unbiased Teacher~\cite{unbiasedTeacher}$^*$ & 20.75$\pm$0.12 & 24.30$\pm$0.07 & 28.27$\pm$0.11 & 31.50$\pm$0.10 & 40.2 $\xrightarrow{+1.1}$ 41.3 \\
    Soft Teacher~\cite{softTeacher}$^\dagger$& 20.46$\pm$0.39 & - & 30.74$\pm$0.08 & 34.04$\pm$0.14 & 40.9 $\xrightarrow{+3.6}$ 44.5 \\

    LabelMatching~\cite{LabelMatching}$^*$ & \textbf{25.81$\pm$0.28} & - & \underline{32.70$\pm$0.18} &35.49$\pm$0.17 & 40.3 $\xrightarrow{+5.0}$ 45.3\\
    PseCo~\cite{pseco}$^\dagger$& 22.43$\pm$0.36 & {27.77$\pm$0.18} & {32.50$\pm$0.08} & \underline{36.06$\pm$0.24} & 41.0 $\xrightarrow{+5.1}$ \textbf{46.1} \\
    DTG-SSOD~\cite{li2022DTG}$^\dagger$ & 21.27$\pm$0.12 & {26.84}$\pm$0.25 & {31.90}$\pm$0.08 & {35.92}$\pm$0.26 & 40.9 $\xrightarrow{{+4.8}}$ \underline{45.7} \\
    
    Unbiased   Teacher v2~\cite{Liu_2022_ubtv2}$^*$ & \underline{25.40$\pm$0.36} & \underline{28.37$\pm$0.03} & 31.85$\pm$0.09 & 35.08$\pm$0.02 & 40.9 $\xrightarrow{{+3.9}}$ {44.8} \\

    MixTeacher (Ours)$^\dagger$ & {25.16$\pm$0.26} & \textbf{29.11$\pm$0.21} & \textbf{34.06$\pm$0.13} & \textbf{36.72$\pm$0.16} & 40.9 $\xrightarrow{+4.8}$ \underline{45.7} \\
     \bottomrule
    \end{tabular}
    }
    \vspace{-.5em}
    \caption{Comparison with state-of-the-art methods on COCO benchmark. $\text{AP}_{50:95}$ on \texttt{val2017} set are reported.
    Under the {Partially Labeled} setting, results are the average of all five folds and numbers behind $\pm$ indicate the standard deviation. Under the {Additional} setting, numbers in front of the arrow indicate the supervised baseline. $\dagger$: using labeled/unlabeled batch size 8/32, * indicates 32/32, and rest of the results using batch size 8/8.
    Bold indicates the best, while underline indicates the second best.
    }
    \label{tab:sota_coco}
    \vspace{-1em}
\end{table*}

\begin{table}[t]
    \centering
    \resizebox{0.9\linewidth}{!}{
\begin{tabular}{lllll}
\toprule
                            & 1\%   & 2\%   & 5\%   & 10\%  \\ \midrule
Dense Teacher~\cite{denseteacher}           & 22.38 & 27.20 & 33.01 & \textbf{37.13} \\
Unbiased Teacher v2~\cite{Liu_2022_ubtv2}       & 22.71 & 26.03 & 30.08 & 32.61 \\
MixTeacher (Ours)         & \textbf{23.83} & \textbf{27.88} & \textbf{33.42} & {36.95} \\ \bottomrule
\end{tabular}
}
\vspace{-.5em}
\caption{Experimental results on the COCO Partially Labeled with FCOS~\cite{fcos}. PLM is not used for our method in this setting.}
\label{tab:fcos}
\vspace{-1em}
\end{table}

\subsection{Implementation Details}
Following the mainstream choice of the community, we adopt Faster-RCNN~\cite{faster} with FPN~\cite{fpn} and ResNet-50~\cite{resnet} as the detection model.
The proposed method is implemented based on the well-known SoftTeacher~\cite{softTeacher}, and we reuse its augmentation and training hyperparameters without any modification for a fair comparison.
All models are trained on 8 GPUs with a base learning rate 0.01. Details of each setting are as follows: In COCO Partially Labeled setting, all models are trained for 180k steps with 1 labeled image and 4 unlabeled images per GPU in each step. The learning rate is decreased by 0.1 at 120k and 160k steps.
In COCO Additional setting, the model is trained for 720k iterations with 4 labeled images and 4 unlabeled images per GPU in each step. The learning rate is decreased by 0.1 at 480k and 680k steps.
In Pascal VOC settings, the model is trained for 40k steps with a constant learning rate with 2 labeled and 2 unlabeled images per GPU in each step.

The weight of unlabeled loss $\alpha$ is set to 4.0 for the COCO Partially Labeled setting, and 2.0 for other settings considering the proportion of labeled and unlabeled images under different settings. Other hyper-parameters are the same for all settings, \ie $T=3.0$, $\tau_h=0.9$, $\tau_l=0.7$, $\delta=0.1$.

\subsection{Comparison with State-of-the-art}

We compare the proposed MixTeacher with the supervised baseline and several recent methods on MS~COCO and Pascal~VOC benchmarks. For a fair comparison, we follow the experimental settings with Soft Teacher~\cite{softTeacher}. 
Without loss of generality, some known tricks which can improve the results in most cases are not used. For instance, Unbiased Teacher~\cite{unbiasedTeacher} and PseCo~\cite{pseco} adopt Focal Loss~\cite{focal} to handle the class imbalance problem. Some methods~\cite{LabelMatching, unbiasedTeacher} adopt a larger batch size for unlabeled data. Besides, it is worth mention that PseCo~\cite{pseco} changes the output levels of FPN from P2-P6 to P3-P7 in testing.

\vspace{.5em}
\noindent \textbf{MS COCO.} We first evaluate the proposed method on MS COCO in~\cref{tab:sota_coco}. Under the Partially Labeled setting, the statistical results over five folds for four different labeled ratios are reported. Our method achieves more than 12\% mAP improvements against the supervised baseline for the settings with less than 10\% labeled data and demonstrates superiority in most of the labeled ratios. 
For the case of 2\% and 5\% labeling ratio, our method yields 29.11 and 34.06 mAP, which is around 1.5 mAP higher than the previous best method PseCo~\cite{pseco}. When the labeled data is extremely scarce, i.e., 1\% labeling ratio, there are only tens labeled images for some tail categories. Our method without any class-specific process still reaches comparable performance to the previous best method LabelMatching~\cite{LabelMatching}, which estimates the class distribution and tunes thresholds adaptively. Under the Additional setting, our method is competitive to the state-of-the-art PseCo~\cite{pseco}. It demonstrates the effectiveness of our method even with adequate labeled data. Besides, we conduct an experiment to evaluate our method on the anchor-free detector FCOS~\cite{fcos} with ResNet-50 backbone. In this case, only the mixed scale teacher is used. The results in \cref{tab:fcos} show that the anchor-free detector can also benefit from the proposed mixed scale teacher. 

\begin{table}[t]
  \centering
  \resizebox{0.82\columnwidth}{!}{
    \begin{tabular}{@{}lll@{}}

    \toprule
     & $\text{AP}_{50}$ & $\text{AP}_{50:95}$ \\
    \midrule
    Supervised \cite{unbiasedTeacher} & 72.63 & 42.13 \\
    \hline
    STAC \cite{STAC} & 77.45 \small{(\textcolor{blue}{$+4.82$})} & 44.64 \small{(\textcolor{blue}{$+2.51$})} \\

    Humble Teacher~\cite{humbleTeacher} & 80.94 \small{(\textcolor{blue}{$+8.31$})} & 53.04  \small{(\textcolor{blue}{$+10.91$})}\\
    Rethinking Pse~\cite{RethinkingPse} & 79.00 \small{(\textcolor{blue}{$+6.37$})} & 54.60  \small{(\textcolor{blue}{$+12.47$})}\\
    LabelMatching~\cite{LabelMatching} & 85.48 \small{(\textcolor{blue}{$+12.85$})} & 55.11 \small{(\textcolor{blue}{$+12.98$})} \\
    MixTeacher (Ours)   & \textbf{85.85}  \small{(\textcolor{blue}{$+13.22$})} &  \textbf{56.25}  \small{(\textcolor{blue}{$+14.12$})}\\
    \bottomrule
  \end{tabular}}
  \vspace{-.5em}
  \caption{Experimental results on the {VOC Additional} setting.}
  \vspace{-1em}
  \label{tab:voc_add}
\end{table}

\begin{table}[t]
  \centering
  \resizebox{0.82\columnwidth}{!}{
    \begin{tabular}{@{}llll@{}}

    \toprule
     & \multicolumn{1}{c}{$\text{AP}_{50}$} & \multicolumn{1}{c}{$\text{AP}_{50:95}$} \\
    \midrule
    Supervised \cite{unbiasedTeacher} & 72.63 & 42.13 \\
    \hline
    STAC \cite{STAC} & 79.08 \small{(\textcolor{blue}{$+6.45$})} & 46.01 \small{(\textcolor{blue}{$+3.88$})} \\

    Humble Teacher~\cite{humbleTeacher} & 81.29 \small{(\textcolor{blue}{$+8.66$})}& 54.41 \small{(\textcolor{blue}{$+12.28$})}\\
    Rethinking Pse~\cite{RethinkingPse} & 79.60 \small{(\textcolor{blue}{$+6.79$})}& 56.10 \small{(\textcolor{blue}{$+13.97$})}\\
    LabelMatching$^\dagger$~\cite{LabelMatching} & 85.81 \small{(\textcolor{blue}{$+13.18$})} &  55.50 \small{(\textcolor{blue}{$+13.37$})} \\
    MixTeacher (Ours)   &  \textbf{86.58} \small{(\textcolor{blue}{$+13.95$})} & \textbf{56.83} \small{(\textcolor{blue}{$+14.70$})} \\
    \bottomrule
  \end{tabular}}
  \vspace{-.5em}
  \caption{Experimental results on the {VOC Mixture} setting.}
  \vspace{-1.5em}
  \label{tab:voc_mix}
\end{table}

\vspace{.5em}
\noindent \textbf{Pascal VOC.} We also evaluate our method under two VOC settings. The results are shown in~\cref{tab:voc_add} and~\cref{tab:voc_mix}. $\dagger$ denotes that we report the results for the method with their official implementation. The results demonstrate that our method consistently reaches the best performance under two settings. Similar to the conclusion of COCO Additional setting, the results indicate the model can benefit from our method even if there is already sufficient labeled data.

\begin{table}[t]
\centering
\resizebox{0.98\columnwidth}{!}{
\begin{tabular}{@{}ccc|cc|c|cc@{}}
\toprule
\multicolumn{3}{c|}{Feature Scales} & \multirow{2}{*}{MST} & \multirow{2}{*}{PLM} & \multirow{2}{*}{mAP} & \multirow{2}{*}{$\text{AP}_{50}$} & \multirow{2}{*}{$\text{AP}_{75}$} \\ 
$\mathbb{P}^+$ & $\mathbb{P}^-$ & $\mathbb{P}^\times$ & & &  &  &  \\ \midrule
   &    &    &  &                              & 26.8  &  45.1      & 28.4  \\ \midrule
\cmark  &    &    &  &                         & 33.9 (+7.1) &  54.0      & 37.0  \\
\cmark  & \cmark  &    & &                     & 34.7 (+7.9) &  54.7      & 37.8  \\
\cmark  & \cmark  &  \cmark & &                & 34.4 (+7.4) &  54.2      & 37.2  \\
\cmark  & \cmark  &  \cmark  & \cmark &        & 36.2 (+9.4) &  56.5      & 39.5  \\
\cmark  & \cmark  &  \cmark  & \cmark & \cmark & \textbf{36.7 (+9.9)} &  \textbf{57.0}      & \textbf{39.7}  \\ \bottomrule
\end{tabular}
}
\vspace{-0.5em}
\caption{Analysis of various components of proposed approach. MST indicates generating pseudo labels from the mixed scale feature pyramid. PLM indicates promising labels mining. }
\label{tab:ablation}
\vspace{-1.5em}
\end{table}

\subsection{Ablation Study}

In this part, we conduct experiments under the COCO Partially Labeled setting to analyze and validate our method in detail. All the experiments in this section are conducted on a single data fold with 10\% labeling ratio.

\vspace{.3em}
\noindent \textbf{Effect of each component.} We validate the effectiveness of each component step by step, and the results are shown in~\cref{tab:ablation}. The model starts from 26.8 mAP when using the labeled data only. After using the regular scale $\mathbb{P}^+$ of unlabeled data, the semi-supervised baseline reaches 33.9 mAP immediately. Furthermore, leveraging additional down-sampled scale features $\mathbb{P}^-$, and with supervision from the regular scale of teacher network, the model achieves another +0.8 mAP improvement. However, there are no gains of introducing a mixed scale features $\mathbb{P}^\times$, but still with the supervision from the regular scale makes the results worse. Instead, with the guidance of pseudo labels from mixed scale, the mAP boosts to 36.2. Finally, our method equips mixed scale pseudo labels and promising labels mining reaches the best performance, 36.7 mAP.

\vspace{.3em}
\noindent \textbf{Comparison with other multi-view methods.} Scale variation across object instances is a key challenge in semi-supervised object detection, and some pioneer works~\cite{sed, pseco} have introduced an additional down-sampled view to improve the model handle the scale issue.~\cref{tab:mv_method} shows the performance and the training time for each iteration of different multi-view methods integrated into the naive implement of Soft Teacher~\cite{softTeacher}. SCR indicates Scale Consistency Regularization in SED~\cite{sed}. MSIL indicates Multi-view Scale-Invariant Learning in PseCo~\cite{pseco}. we report the results of MST$^\ddagger$ training with randomly dropping a path from large scale and mixed scale for the loss of student network, \ie only the 0.5$\times$ scale and one of 1$\times$ scales are used to keep a comparable training time with other multi-scale methods (1.03 \vs 0.97 \vs 0.93 sec/iter). The results show that all methods with additional view improve the performance compared with the single view baseline. Our method presents a significant advantage of more than 1.3 mAP gains compared with previous consistency learning methods. Randomly dropping a path in training saves 0.19 second every iteration and reaches comparable results. 

\begin{table}[t]
\centering
\begin{minipage}[t]{0.5\textwidth}
  \centering
  \resizebox{0.75\textwidth}{!}{
\begin{tabular}{@{}p{2.8cm}|c|cc|c@{}}
\toprule
                    & mAP & $\text{AP}_{50}$ & $\text{AP}_{75}$ & sec/iter \\ \midrule
Baseline            &   33.9 &	54.0 &	37.0 &	0.75   \\ \midrule
SCR~\cite{sed}      &   34.6 &	54.6 &	37.9 &	0.97 \\
MSIL~\cite{pseco}   &   34.9 &	55.1 &	37.6 &	\textbf{0.94} \\
MST$^\ddagger$ (Ours)     &   36.0 &	56.3 &	\textbf{39.5} &  1.03 \\
MST (Ours)          &   \textbf{36.2} &	\textbf{56.5} &	\textbf{39.5} &  1.22 \\ \bottomrule
\end{tabular}}
\vspace{-.5em}
\caption{Comparison with other multi-view methods.}
\label{tab:mv_method}
\end{minipage}

\vspace{0.5em}
\centering
\begin{minipage}[t]{0.5\textwidth}
\centering
  \resizebox{0.75\textwidth}{!}{
\begin{tabular}{@{}p{3.2cm}|c|cc@{}}
\toprule
                      & mAP & $\text{AP}_{50}$ & $\text{AP}_{75}$ \\ \midrule
Baseline                        & 34.7  &  54.7      & 37.8       \\ \midrule
CONV-ADD                 & 35.1 (+0.4) &  55.0      & 38.5       \\
CAT-CONV           & 35.2 (+0.5)  &  55.2      & 38.3       \\

GAP-MLP (Ours)                & \textbf{36.2 (+1.5)} &  \textbf{56.5}      & \textbf{39.5}       \\ \bottomrule
\end{tabular}}
\vspace{-.5em}
\caption{Comparison of feature fusion approaches.}
\label{tab:fusion_arch}
\end{minipage}

\vspace{0.5em}
\centering
\begin{minipage}[t]{0.5\textwidth}
  \centering
  \resizebox{0.75\textwidth}{!}{
\begin{tabular}{@{}p{2cm}|c|ccc|c@{}}
\toprule
                    & mAP & $\text{AP}_s$ & $\text{AP}_m$ & $\text{AP}_l$ & FPS \\ \midrule
Baseline            &     36.5 &	21.8 &	39.2 &	48.6  &  33.4 \\ \midrule
Test on $\mathbb{P}^-$      &   33.2 &	14.6 &	36.1 &	50.0  &  \textbf{37.1}   \\
Test on $\mathbb{P}^+$      &   36.7 &	\textbf{21.8} &	39.2 &	48.6  &  33.4  \\
Test on $\mathbb{P}^\times$ &   \textbf{37.5} &	\textbf{21.8} &	\textbf{40.1} &	\textbf{51.1}   &  27.0   \\ \bottomrule
\end{tabular}}

\vspace{-.1em}
\caption{Performance of different scales testing.}
\label{tab:test_scale}
\end{minipage}
\vspace{-1.5em}
\end{table}

\begin{figure*}[t]
	\centering
	\includegraphics[height=0.8\columnwidth, width=0.95\textwidth]{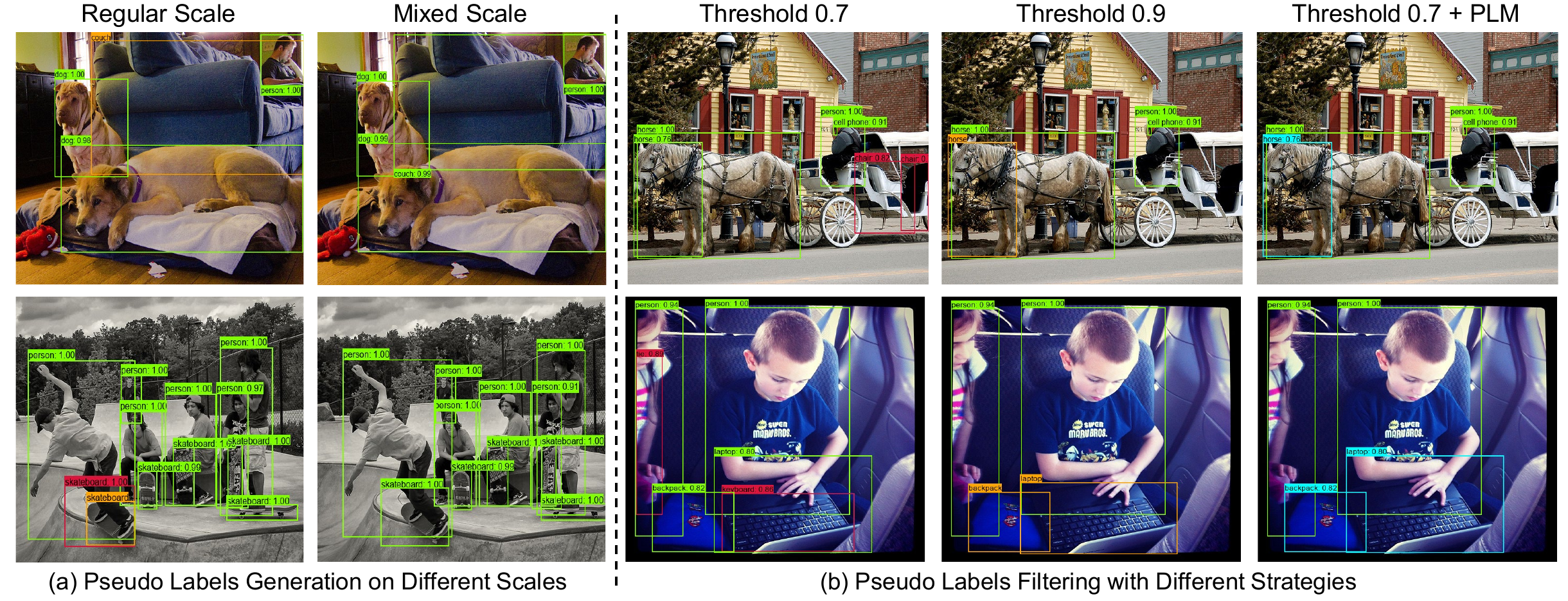}
    \vspace{-.5em}
	\caption{Qualitative visualization for the components in MixTeacher. (a) Comparison of pseudo labels generated from the regular scale and the mixed scale feature pyramids. (b) Comparison of pseudo labels under different score thresholds and our promising label mining results. The \textcolor{green}{green} boxes denote True Positives. The \textcolor{red}{red} boxes highlight the False Positives, and \textcolor{orange}{orange} boxes denote the False Negatives. Besides, the mined labels are highlight with a \textcolor{cyan}{cyan} box.}
	\vspace{-.5em}
	\label{fig:visualization}
\end{figure*}

\begin{table*}[t]
    \centering
    \label{tab:ablation_param}
    \begin{subtable}[t]{0.37\textwidth}
    \centering
    
    \renewcommand{\arraystretch}{1.22}
    \resizebox{0.9\textwidth}{!}{
     \begin{tabular}{c|c|c|c|c|c}
     \toprule
       $T$ & 0.1 & 0.3 & 1.0 & 3.0 & 10.0  \\
     \midrule
     mAP       & 35.9 & 35.9 & 36.1 & \textbf{36.2} & 36.1 \\
     AP$_{50}$ & 56.4 & 56.3 & 56.3 & \textbf{56.5}  & \textbf{56.5}\\
     AP$_{75}$ & 39.3 & 39.2 & 39.4 & \textbf{39.5} & 39.2\\ \bottomrule
    \end{tabular}
    }
    \caption{Study on {temperature} $T$.}
    \end{subtable} \hfill
    \begin{subtable}[t]{0.26\textwidth}
    \centering
    \renewcommand{\arraystretch}{1.19}
    \resizebox{0.9\textwidth}{!}{
    \begin{tabular}{c|c|cc}
    \toprule
     $\tau_l$ & mAP & AP$_{50}$ & AP$_{75}$  \\ \midrule
        0.6   &  33.8 & 54.2 & 37.1  \\
        0.7   &  \textbf{36.7} & \textbf{57.0} & \textbf{39.8} \\
        0.8   &  36.3 & 56.8 & 39.2 \\ \bottomrule
    \end{tabular}
    }
    \caption{Study on {lower threshold $\tau_l$}}
    \end{subtable} \hfill
    \begin{subtable}[t]{0.34\textwidth}
    \centering
    \renewcommand{\arraystretch}{1.22}
     \resizebox{0.9\textwidth}{!}{
     \begin{tabular}{c|c|c|cc}
     \toprule
     $\delta$ & AvgBox & mAP & AP$_{50}$ & AP$_{75}$  \\ \midrule
     0.0 & 3.87 & 36.2 & 56.5 & 39.5 \\
     0.1 & 4.15 & \textbf{36.7} & \textbf{57.0} & \textbf{39.8} \\
     0.2 & 3.91 & 36.2 & 56.7 & 39.5 \\ \bottomrule
     \end{tabular} 
     }
     \caption{Study on {promotion threshold $\delta$}.}
    \end{subtable}
    \vspace{-.6em}
    \caption{Comparison of different hyper-parameters for the proposed MixTeacher.}
    \vspace{-1.2em}
    \label{tab:hyperparam}
\end{table*}

\noindent \textbf{Comparison of feature fusion approaches.} we compare the effects of different feature fusion approaches to build mixed scale features. We evaluate the performance of three simple fusion architectures, \ie ``CONV-ADD" denotes that employ two 3$\times$3 convolution layer to align features for regular scale and down-sampled scale, followed with an element-wise addition, ``CAT-CONV" denotes that concatenated by channel and then apply convolution to reduce channels. As shown in~\cref{tab:fusion_arch}, compared with the baseline without fusion features, all three fusion methods obtain gains in mAP. Among them, our method that builds the mixed scale feature as a weighted summation of regular and down-sampled scales achieves the best performance.

\noindent \textbf{Performance of different scales testing.} In the proposed method, each feature pyramid in three scales is capable to detect objects. The inference results with different feature scales are useful to guide the design of the strategy for pseudo label generation, and can also be used as a separate detector. Table~\ref{tab:test_scale} shows the performance of a model with testing on different feature scales. We also report the average inference speed on a single V100 GPU. Notice that the pipeline and architecture are exactly the same as the vanilla faster R-CNN when tested on the regular or down-sampled scale. However, additional computation and parameters are required to build the feature pyramid when tested on mixed scale. The results show that the down-sampled scale makes the worst mAP and falls behind the regular scale one on $\text{AP}_s$ especially, but it reaches the best $\text{AP}_l$ for large objects. It suggests that previous methods~\cite{sed, pseco} generating pseudo labels from the regular scale is not appropriate for large objects. On the other hand, the mixed scale which adaptive select scales achieves competitive results on all objects and it is more suitable for pseudo labels generation.

\vspace{.3em}
\noindent \textbf{Choice of hyper-parameters.} In order to analyze the sensitivity of some key hyper-parameters, we investigate the influence of different temperature $T$ in building mixed scale feature pyramid, the lower score threshold $\tau$ for promising labels mining, and the score promotion threshold $\delta$ between scales. As shown in~\cref{tab:hyperparam}, the temperature $T$ is set to 3.0 for better results, which encourages the feature space of mixed scale to be more likely to one of the source scales. The lower threshold $\tau_l$ is set to 0.7, which controls the number of candidates for mining. The promotion threshold $\delta$ is set to 0.1, which mines 0.28 boxes for a image in average and brings 0.5 gains in mAP.

\subsection{Qualitative Visualization}
We show qualitative results to demonstrate the quality of pseudo labels more intuitively. As shown in~\cref{fig:visualization} (a), the pseudo labels generated from the proposed mixed scale features are more accurate than the regular scale. \cref{fig:visualization} (b) shows that there are many false positives when threshold=0.7 and false negatives when threshold=0.9, our promising label mining method alleviates these problems.

\section{Conclusion and Limitation}
In this work, we delve into the scale variation problem in semi-supervised object detection, and propose a novel framework by
introducing a mixed scale teacher to improve the pseudo labels generation and scale invariant learning. In addition,
benefiting from better predictions from mixed scale features, we propose to mine pseudo labels with the score promotion of predictions across scales.  Extensive experiments on MS COCO and Pascal VOC benchmarks under various semi-supervised settings demonstrate that our method achieves new state-of-the-art performance. While we have shown the superiority of MixTeacher, the method is built on an old-fashion detector with the simplest FPN and naive label assignment strategy. Whether the scale variation problem in SSOD can be addressed with more advanced FPN architectures or label assignment methods is unclear, which is an interesting future work.
\newpage

{\small
\bibliographystyle{ieee_fullname}
\bibliography{ref}

\begin{thebibliography}{10}\itemsep=-1pt

\bibitem{bachman2014learning}
Philip Bachman, Ouais Alsharif, and Doina Precup.
\newblock Learning with pseudo-ensembles.
\newblock {\em Advances in neural information processing systems}, 27, 2014.

\bibitem{mixmatch}
David Berthelot, Nicholas Carlini, Ian Goodfellow, Nicolas Papernot, Avital
  Oliver, and Colin~A Raffel.
\newblock Mixmatch: A holistic approach to semi-supervised learning.
\newblock {\em Advances in Neural Information Processing Systems}, 32, 2019.

\bibitem{LabelMatching}
Binbin Chen, Weijie Chen, Shicai Yang, Yunyi Xuan, Jie Song, Di Xie, Shiliang
  Pu, Mingli Song, and Yueting Zhuang.
\newblock Label matching semi-supervised object detection.
\newblock In {\em Proceedings of the IEEE/CVF Conference on Computer Vision and
  Pattern Recognition}, pages 14381--14390, 2022.

\bibitem{VCL}
Changrui Chen, Kurt Debattista, and Jungong Han.
\newblock Semi-supervised object detection via virtual category learning.
\newblock In {\em European Conference on Computer Vision}, 2022.

\bibitem{cityscapes}
Marius Cordts, Mohamed Omran, Sebastian Ramos, Timo Rehfeld, Markus Enzweiler,
  Rodrigo Benenson, Uwe Franke, Stefan Roth, and Bernt Schiele.
\newblock The cityscapes dataset for semantic urban scene understanding.
\newblock In {\em Proceedings of the IEEE/CVF Conference on Computer Vision and
  Pattern Recognition}, 2016.

\bibitem{pascalvoc}
M. Everingham, L. Van~Gool, C.~K.~I. Williams, J. Winn, and A. Zisserman.
\newblock The pascal visual object classes (voc) challenge.
\newblock {\em IJCV}, 88(2):303--338, June 2010.

\bibitem{tood}
Chengjian Feng, Yujie Zhong, Yu Gao, Matthew~R Scott, and Weilin Huang.
\newblock Tood: Task-aligned one-stage object detection.
\newblock In {\em Proceedings of the IEEE/CVF International Conference on
  Computer Vision (ICCV)}, pages 3490--3499. IEEE Computer Society, 2021.

\bibitem{MutualSup}
Ziteng Gao, Limin Wang, and Gangshan Wu.
\newblock Mutual supervision for dense object detection.
\newblock In {\em Proceedings of the IEEE/CVF International Conference on
  Computer Vision (ICCV)}, pages 3641--3650, 2021.

\bibitem{ota}
Zheng Ge, Songtao Liu, Zeming Li, Osamu Yoshie, and Jian Sun.
\newblock Ota: Optimal transport assignment for object detection.
\newblock In {\em Proceedings of the IEEE/CVF Conference on Computer Vision and
  Pattern Recognition}, pages 303--312, 2021.

\bibitem{sed}
Qiushan Guo, Yao Mu, Jianyu Chen, Tianqi Wang, Yizhou Yu, and Ping Luo.
\newblock Scale-equivalent distillation for semi-supervised object detection.
\newblock In {\em Proceedings of the IEEE/CVF Conference on Computer Vision and
  Pattern Recognition}, 2022.

\bibitem{resnet}
Kaiming He, Xiangyu Zhang, Shaoqing Ren, and Jian Sun.
\newblock Deep residual learning for image recognition.
\newblock In {\em Proceedings of the IEEE conference on computer vision and
  pattern recognition}, pages 770--778, 2016.

\bibitem{SENet}
Jie Hu, Li Shen, and Gang Sun.
\newblock Squeeze-and-excitation networks.
\newblock In {\em Proceedings of the IEEE conference on computer vision and
  pattern recognition}, pages 7132--7141, 2018.

\bibitem{CSD}
Jisoo Jeong, Seungeui Lee, Jeesoo Kim, and Nojun Kwak.
\newblock Consistency-based semi-supervised learning for object detection.
\newblock {\em Advances in neural information processing systems}, 32, 2019.

\bibitem{temporalensembling}
Samuli Laine and Timo Aila.
\newblock Temporal ensembling for semi-supervised learning.
\newblock In {\em International Conference on Learning Representations}, 2017.

\bibitem{pseudo-label1}
Dong-Hyun Lee et~al.
\newblock Pseudo-label: The simple and efficient semi-supervised learning
  method for deep neural networks.
\newblock In {\em Workshop on challenges in representation learning, ICML},
  page 896, 2013.

\bibitem{li2022DTG}
Gang Li, Xiang Li, Yujie Wang, Yichao Wu, Ding Liang, and Shanshan Zhang.
\newblock Dtg-ssod: Dense teacher guidance for semi-supervised object
  detection.
\newblock In {\em Advances in neural information processing systems}, 2022.

\bibitem{pseco}
Gang Li, Xiang Li, Yujie Wang, Yichao Wu, Ding Liang, and Shanshan Zhang.
\newblock {PseCo}: {Pseudo} {Labeling} and {Consistency} {Training} for
  {Semi}-{Supervised} {Object} {Detection}.
\newblock In {\em European Conference on Computer Vision}, 2022.

\bibitem{RethinkingPse}
Hengduo Li, Zuxuan Wu, Abhinav Shrivastava, and Larry~S Davis.
\newblock Rethinking pseudo labels for semi-supervised object detection.
\newblock {\em arXiv preprint arXiv:2106.00168}, 2021.

\bibitem{dwla}
Shuai Li, Chenhang He, Ruihuang Li, and Lei Zhang.
\newblock A dual weighting label assignment scheme for object detection.
\newblock In {\em Proceedings of the IEEE/CVF Conference on Computer Vision and
  Pattern Recognition}, pages 9387--9396, 2022.

\bibitem{gfl}
Xiang Li, Wenhai Wang, Lijun Wu, Shuo Chen, Xiaolin Hu, Jun Li, Jinhui Tang,
  and Jian Yang.
\newblock Generalized focal loss: Learning qualified and distributed bounding
  boxes for dense object detection.
\newblock {\em Advances in Neural Information Processing Systems},
  33:21002--21012, 2020.

\bibitem{trident}
Yanghao Li, Yuntao Chen, Naiyan Wang, and Zhaoxiang Zhang.
\newblock Scale-aware trident networks for object detection.
\newblock In {\em Proceedings of the IEEE/CVF International Conference on
  Computer Vision (ICCV)}, pages 6054--6063, 2019.

\bibitem{fpn}
Tsung-Yi Lin, Piotr Doll{\'a}r, Ross Girshick, Kaiming He, Bharath Hariharan,
  and Serge Belongie.
\newblock Feature pyramid networks for object detection.
\newblock In {\em Proceedings of the IEEE conference on computer vision and
  pattern recognition}, pages 2117--2125, 2017.

\bibitem{focal}
Tsung-Yi Lin, Priya Goyal, Ross Girshick, Kaiming He, and Piotr Doll{\'a}r.
\newblock Focal loss for dense object detection.
\newblock In {\em Proceedings of the IEEE international conference on computer
  vision}, pages 2980--2988, 2017.

\bibitem{coco}
Tsung-Yi Lin, Michael Maire, Serge Belongie, James Hays, Pietro Perona, Deva
  Ramanan, Piotr Doll{\'a}r, and C~Lawrence Zitnick.
\newblock Microsoft coco: Common objects in context.
\newblock In {\em European conference on computer vision}, pages 740--755.
  Springer, 2014.

\bibitem{unbiasedTeacher}
Yen-Cheng Liu, Chih-Yao Ma, Zijian He, Chia-Wen Kuo, Kan Chen, Peizhao Zhang,
  Bichen Wu, Zsolt Kira, and Peter Vajda.
\newblock Unbiased teacher for semi-supervised object detection.
\newblock {\em arXiv preprint arXiv:2102.09480}, 2021.

\bibitem{Liu_2022_ubtv2}
Yen-Cheng Liu, Chih-Yao Ma, and Zsolt Kira.
\newblock Unbiased teacher v2: Semi-supervised object detection for anchor-free
  and anchor-based detectors.
\newblock In {\em Proceedings of the IEEE/CVF Conference on Computer Vision and
  Pattern Recognition}, pages 9819--9828, June 2022.

\bibitem{miyato2018virtual}
Takeru Miyato, Shin-ichi Maeda, Masanori Koyama, and Shin Ishii.
\newblock Virtual adversarial training: a regularization method for supervised
  and semi-supervised learning.
\newblock {\em IEEE Transactions on Pattern Analysis and Machine Intelligence},
  41(8):1979--1993, 2018.

\bibitem{ouali2020semi}
Yassine Ouali, C{\'e}line Hudelot, and Myriam Tami.
\newblock Semi-supervised semantic segmentation with cross-consistency
  training.
\newblock In {\em Proceedings of the IEEE/CVF Conference on Computer Vision and
  Pattern Recognition}, pages 12674--12684, 2020.

\bibitem{faster}
Shaoqing Ren, Kaiming He, Ross Girshick, and Jian Sun.
\newblock Faster r-cnn: Towards real-time object detection with region proposal
  networks.
\newblock {\em Advances in neural information processing systems}, 28, 2015.

\bibitem{giou}
Hamid Rezatofighi, Nathan Tsoi, JunYoung Gwak, Amir Sadeghian, Ian Reid, and
  Silvio Savarese.
\newblock Generalized intersection over union: A metric and a loss for bounding
  box regression.
\newblock In {\em Proceedings of the IEEE/CVF conference on computer vision and
  pattern recognition}, pages 658--666, 2019.

\bibitem{sajjadi2016regularization}
Mehdi Sajjadi, Mehran Javanmardi, and Tolga Tasdizen.
\newblock Regularization with stochastic transformations and perturbations for
  deep semi-supervised learning.
\newblock {\em Advances in neural information processing systems}, 29, 2016.

\bibitem{objects365}
Shuai Shao, Zeming Li, Tianyuan Zhang, Chao Peng, Gang Yu, Xiangyu Zhang, Jing
  Li, and Jian Sun.
\newblock Objects365: A large-scale, high-quality dataset for object detection.
\newblock In {\em Proceedings of the IEEE/CVF International Conference on
  Computer Vision (ICCV)}, October 2019.

\bibitem{snip}
Bharat Singh and Larry~S Davis.
\newblock An analysis of scale invariance in object detection snip.
\newblock In {\em Proceedings of the IEEE/CVF Conference on Computer Vision and
  Pattern Recognition}, pages 3578--3587, 2018.

\bibitem{Fixmatch}
Kihyuk Sohn, David Berthelot, Nicholas Carlini, Zizhao Zhang, Han Zhang,
  Colin~A Raffel, Ekin~Dogus Cubuk, Alexey Kurakin, and Chun-Liang Li.
\newblock Fixmatch: Simplifying semi-supervised learning with consistency and
  confidence.
\newblock {\em Advances in Neural Information Processing Systems}, 33:596--608,
  2020.

\bibitem{STAC}
Kihyuk Sohn, Zizhao Zhang, Chun-Liang Li, Han Zhang, Chen-Yu Lee, and Tomas
  Pfister.
\newblock A simple semi-supervised learning framework for object detection.
\newblock {\em arXiv preprint arXiv:2005.04757}, 2020.

\bibitem{humbleTeacher}
Yihe Tang, Weifeng Chen, Yijun Luo, and Yuting Zhang.
\newblock Humble teachers teach better students for semi-supervised object
  detection.
\newblock In {\em Proceedings of the IEEE/CVF Conference on Computer Vision and
  Pattern Recognition}, pages 3132--3141, 2021.

\bibitem{MeanTeacher}
Antti Tarvainen and Harri Valpola.
\newblock Mean teachers are better role models: Weight-averaged consistency
  targets improve semi-supervised deep learning results.
\newblock {\em Advances in neural information processing systems}, 30, 2017.

\bibitem{fcos}
Zhi Tian, Chunhua Shen, Hao Chen, and Tong He.
\newblock Fcos: Fully convolutional one-stage object detection.
\newblock In {\em Proceedings of the IEEE/CVF international conference on
  computer vision}, pages 9627--9636, 2019.

\bibitem{UDA}
Qizhe Xie, Zihang Dai, Eduard Hovy, Thang Luong, and Quoc Le.
\newblock Unsupervised data augmentation for consistency training.
\newblock {\em Advances in Neural Information Processing Systems},
  33:6256--6268, 2020.

\bibitem{softTeacher}
Mengde Xu, Zheng Zhang, Han Hu, Jianfeng Wang, Lijuan Wang, Fangyun Wei, Xiang
  Bai, and Zicheng Liu.
\newblock End-to-end semi-supervised object detection with soft teacher.
\newblock In {\em Proceedings of the IEEE/CVF International Conference on
  Computer Vision (ICCV)}, pages 3060--3069, 2021.

\bibitem{ISMT}
Qize Yang, Xihan Wei, Biao Wang, Xian-Sheng Hua, and Lei Zhang.
\newblock Interactive self-training with mean teachers for semi-supervised
  object detection.
\newblock In {\em Proceedings of the IEEE/CVF Conference on Computer Vision and
  Pattern Recognition}, pages 5941--5950, 2021.

\bibitem{polish}
Lei Zhang, Yuxuan Sun, and Wei Wei.
\newblock Mind the gap: Polishing pseudo labels for accurate semi-supervised
  object detection.
\newblock {\em arXiv preprint arXiv:2207.08185}, 2022.

\bibitem{atss}
Shifeng Zhang, Cheng Chi, Yongqiang Yao, Zhen Lei, and Stan~Z Li.
\newblock Bridging the gap between anchor-based and anchor-free detection via
  adaptive training sample selection.
\newblock In {\em Proceedings of the IEEE/CVF Conference on Computer Vision and
  Pattern Recognition}, pages 9759--9768, 2020.

\bibitem{freeanchor}
Xiaosong Zhang, Fang Wan, Chang Liu, Rongrong Ji, and Qixiang Ye.
\newblock Freeanchor: Learning to match anchors for visual object detection.
\newblock {\em Advances in neural information processing systems}, 32, 2019.

\bibitem{zhao2020sess}
Na Zhao, Tat-Seng Chua, and Gim~Hee Lee.
\newblock Sess: Self-ensembling semi-supervised 3d object detection.
\newblock In {\em Proceedings of the IEEE/CVF Conference on Computer Vision and
  Pattern Recognition}, pages 11079--11087, 2020.

\bibitem{denseteacher}
Hongyu Zhou, Zheng Ge, Songtao Liu, Weixin Mao, Zeming Li, Haiyan Yu, and Jian
  Sun.
\newblock Dense teacher: Dense pseudo-labels for semi-supervised object
  detection.
\newblock In {\em European Conference on Computer Vision}, page 35–50,
  Berlin, Heidelberg, 2022. Springer-Verlag.

\bibitem{instantTeaching}
Qiang Zhou, Chaohui Yu, Zhibin Wang, Qi Qian, and Hao Li.
\newblock Instant-teaching: An end-to-end semi-supervised object detection
  framework.
\newblock In {\em Proceedings of the IEEE/CVF Conference on Computer Vision and
  Pattern Recognition}, pages 4081--4090, 2021.

\end{thebibliography}
}

\clearpage
\vspace{1em}
\noindent{\textbf{\LARGE{Supplementary Material}}}

\renewcommand\thesection{\Roman{section}}
\setcounter{section}{0}
\setcounter{table}{0}
\setcounter{equation}{0}
\setcounter{figure}{0}

\section{Experimental Details}

Different SSOD methods may implement with different data augmentation strategies and training hyper-parameters which have a great impact on the performance. As the choice of the majority, our implementation and hyper-parameters are based on MMDetection, with the base model of \texttt{FasterRCNN-R50-FPN}. We implement MixTeacher without any modification on the model design and loss formulation, except for the necessary module and losses introduced by the mixed scale teacher in training (which are dropped in testing). The training hyper-parameters are summarized in~\cref{tab:hyper-param}.

\begin{table}[h]
  \centering
    \resizebox{0.95\columnwidth}{!}{
    
    \begin{tabular}{lccc}
    \toprule
    Training Setting & COCO-Partial & COCO-Additional & VOC \\
    \midrule
    Batch size for labeled data & 8 & 32 & 16 \\
    Batch size for unlabeled data & 32 & 32 & 16 \\
    Learning rate & 0.01 & 0.01 & 0.01 \\
    Learning rate step & (120k, 160k) & (480k, 640k) & - \\
    Iterations & 180k & 720k & 40k \\
    Unsupervised loss weight $\lambda$ & 4.0 & 2.0 & 2.0 \\
    EMA rate & 0.999 & 0.999 & 0.999  \\
    Temperature $T$ & 3 & 3 & 3 \\
    Mine score thresh $\tau_{l}$ & 0.7 & 0.7 & 0.7 \\
    Mine diff thresh $\delta$ & 0.1 & 0.1 & 0.1 \\
    
    Test score threshold & 0.001 & 0.001 & 0.001 \\
    \bottomrule
  \end{tabular}}
  \caption{The summary of training settings for different settings.}
  \label{tab:hyper-param}
\end{table}

\begin{table*}[h]
\centering
    \resizebox{0.8\textwidth}{!}{
\vspace{0em}
\begin{tabular}{@{}c|c|c|c@{}}
\toprule
Augmentation & Labeled Data Aug.     & Unlabeled Strong Aug.   & Unlabeled Weak Aug. \\ \midrule
Scale jitter & short edge $\in$ (0.5,1.5)        & short edge $\in$ (0.5,1.5)       & short edge $\in$ (0.5,1.5)     \\
Horizontal flip & p=0.5 & p=0.5 & p=0.5\\
Solarize jitter   & p=0.25, ratio $\in$ (0,1) & p=0.25, ratio $\in$ (0,1)                & -- \\
Brightness jitter & p=0.25, ratio $\in$ (0,1) & p=0.25, ratio $\in$ (0,1)               & -- \\
Constrast jitter  & p=0.25, ratio $\in$ (0,1) & p=0.25, ratio $\in$ (0,1)                & -- \\
Sharpness jitter  & p=0.25, ratio $\in$ (0,1) & p=0.25, ratio $\in$ (0,1)                & -- \\
Translation       & --                  & p=0.3, translation ratio $\in$ (0,0.1) & -- \\
Rotate            & --                  & p=0.3, angle $\in$ (0,30°)               & -- \\
Shift             & --                  & p=0.3, angle $\in$ (0,30°)               & -- \\
Cutout       & num $\in$ (1,5), ratio $\in$ (0.05,0.2) & num $\in$ (1,5), ratio $\in$ (0.05,0.2) & --                       \\ \bottomrule
\end{tabular}}
\vspace{-.0em}
\caption{The summary of training settings for different datasets and different settings. We follow the practive of Soft Teacher~\cite{softTeacher}, STAC~\cite{STAC}, and FixMatch~\cite{Fixmatch} to adopt different hyper-parameters for labeled data augmentation, and unlabeled strong-weak augmentation.}
\label{tab:augment}
\vspace{0em}
\end{table*}

Note that, as we illustrated in the main manuscript, we adopt the confidence score thresholds $\tau_h=0.9$ and $\tau_l=0.7$ to select and mine pseudo labels for the classification loss of RCNN and the classification and regression losses of RPN. Moreover, we follow the practice in Soft Teacher~\cite{softTeacher} which adopts a different strategy to filter out pseudo labels for the regression loss of RCNN. Concretely, the pseudo labels with a confidence score higher than 0.5 are selected as the candidates, and the candidates with uncertainty lower than 0.02 are selected for RCNN regression. The estimation of uncertainty for the box localization reliability is implemented by jittering each predicted box 10 times as a group of proposals, and computing the standard deviation of the corresponding location predictions for the group of proposals. The offsets of jittering are uniformly sampled from [-6\%, 6\%] of the height or width of the pseudo box candidates. If necessary, please refer to \cite{softTeacher} for the details of this part.

In addition, strong-weak augmentation is commonly used in semi-supersvied learning, we follow previous works to use different augmentations for labeled data, unlabeled images, and pseudo labels generation during training. The details of data augmentation are summarized in \cref{tab:augment}.

\section{Training Efficiency}

\begin{figure*}[t]
	\centering
\vspace{0em}
	\includegraphics[width=0.92\textwidth]{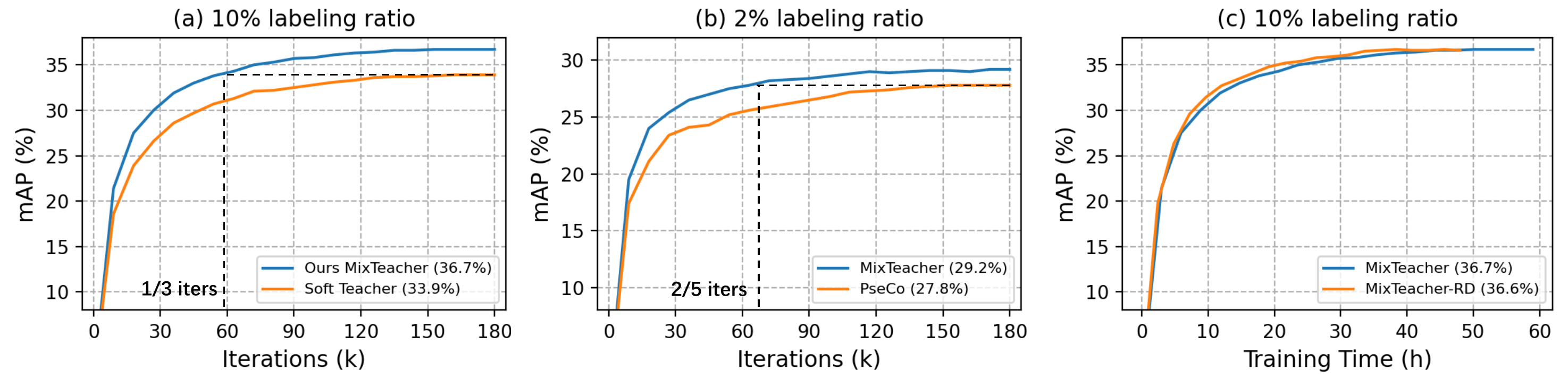}
	\caption{Comparison of model convergence speed in COCO partial labeled setting. (a) Compare MixTeacher against Soft Teacher~\cite{softTeacher} under 10\% labeling ratio. (b) Compare MixTeacher against PseCo~\cite{pseco} under 1\% labeling ratio. (c) Compare MixTeacher against MixTeacher-RD under 10\% labelling ratio, which randomly drops a path from the regular scale and the mixed scale for unlabeled images in every iteration. In legend, the numbers in brackets refer to the final mAP. Performance is evaluated on the teacher model.} 
	\label{fig:train_eff}
\end{figure*}

Since our method brings extra computations for additional scales, the major concern might be the training efficiency of our method. Although we have reported the training speed in Table 5 of the main manuscript, we further investigate the convergence speed for different models. We plot the evaluation results during training for different models in~\cref{fig:train_eff}. Compared with our baseline, \ie the original version of soft teacher, our method obtains a similar result of 33.9 mAP with only 1/3 of total iterations, and finally reaches a significant improvement to 36.7 mAP. Comparing with the most recent method PseCo~\cite{pseco} which also uses an additional down-sampled view but still generates pseudo labels from the regular scale, our method also behaves superiority, for which obtains a comparable result with only 40\% iteration. Furthermore, we conduct a experiment to compare the proposed MixTeacher with a version named MixTeacher-RD that ramdonly drops a $1\times$ scale path for the student model in each iteration of unlabeled data. As shown in~\cref{fig:train_eff} (c), randomly dropping a path can reduce the time consumption of each training iteration, but still reaches a comparable results in the end. More specifically, we report the performance of MixTeacher-RD under all four labelling ratio of COCO partial label settings in \cref{tab:sota_coco}. The results demonstrate that when using a single $1\times$ scale view and a $0.5\times$ scale view as previous multiple views SSOD methods~\cite{sed, pseco}, our method still improves the performance significantly.

\begin{figure}[b]
  \centering
  
  \includegraphics[width=0.95\linewidth]{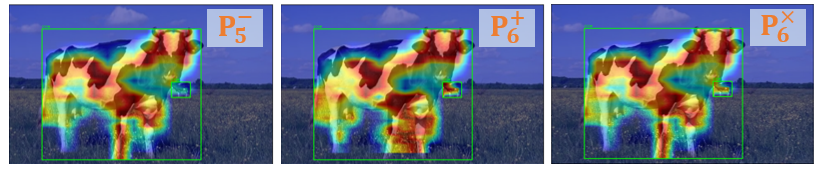}
  
   \caption{EigenCAM visualization for layers in different feature pyramids. $\gamma=0.18$ leads the $P_6^\times$ more similar to $P_5^-$. }
  
   \label{fig:vis}
\end{figure}

\section{Feature Visulization}
 $\gamma$ is derived from channel-wise attention on the regular scale and down-sampled scale features, serving as a weight in the linear combination of these two features. The weighted sum formulation acts as a gate mechanism to select more appropriate feature for each level. We show activation maps in Fig.\ref{fig:vis} for an image with two cows in different sizes. In this case, the 5th level of the down-sampled pyramid shows more accurate for the large cow than the 6th level of the regular scale pyramid. On the other hand, the smaller cow shows a higher response in the regular scale, but it is not appropriate to detect in this level due to its size. Thus, a lower $\gamma$ that tends to use the down-sampled scale is appropriate.

\section{Bells and whistles in SSOD}
In order to avoid the confusion about what makes results better, we follow a quite simple baseline, in which some tricks that known to improve results are not used. For instance, PseCo~\cite{pseco} uses Focal Loss~\cite{focal} to replace the cross entropy loss in the original Faster-RCNN implementation, which has been proven can bring $+0.6$ mAP improvement in their work. Unbiased Teacher~\cite{unbiasedTeacher} adopts a larger batch size than ours, and we tried it on our baseline with getting $+0.3$ mAP gains but increasing the training time. 

Besides, inspired from the progress in fully supervised object detection, such as GIoU loss~\cite{giou}, dynamically hard label assignment~\cite{atss, ota}, and soft label assignment~\cite{gfl, freeanchor}, recent SSOD methods resort to more advanced label assignment strategies~\cite{pseco, LabelMatching}, or more efficient localization loss~\cite{polish}. 
We keep the ordinary implementation to demonstrate the effective of proposed method. We believe it is unnecessary to spend time on trying these components, although they are highly likely to bring better results for MixTeacher, .

\begin{figure}[th]
	\centering
	\includegraphics[width=0.99\columnwidth]{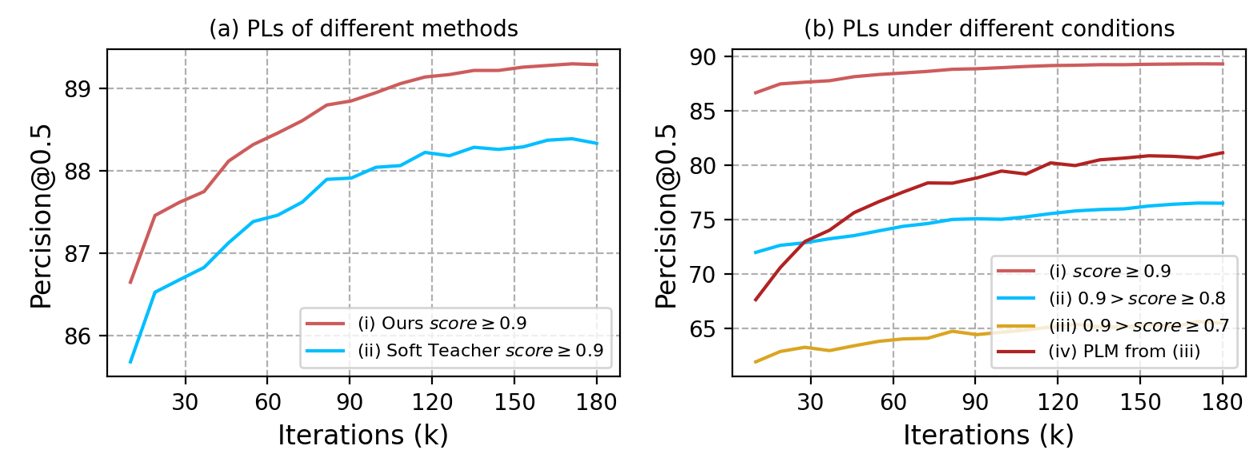}
    
	\caption{Comparison of the quality of pseudo labels during training. (a) Compare MixTeacher against Soft Teacher~\cite{softTeacher} under 10\% labeling ratio. (b) Compare the pseudo labels of MixTeacher under different conditions. The pseudo labels with IoU overlapping the ground truth greater than 0.5 are regarded as true positives} 
	\label{fig:quality}
\end{figure}

\begin{table*}[th]
    \centering
    \resizebox{0.76\linewidth}{!}{
    \renewcommand{\arraystretch}{1.05}
    \begin{tabular}{p{4.cm}|c|c|c|c|c}
    \toprule
      & Unlabeled Data & \multicolumn{4}{c}{COCO Partially Labeled}  \\ 
     & Views Used & 1\% & 2\% & 5\% & 10\% \\ \midrule
    Supervised Baseline & None & 12.15$\pm$0.27 & 16.65$\pm$0.18 & 21.45$\pm$0.16 & 27.10$\pm$0.07  \\ \midrule
    Soft Teacher~\cite{softTeacher}  & $\{1\times\}$ & 20.46$\pm$0.39 & - & 30.74$\pm$0.08 & 34.04$\pm$0.14  \\ 
    SED~\cite{sed}  & $\{1\times, 0.5\times\}$ & - & - & 29.01 & 34.02  \\
    PseCo~\cite{pseco}& $\{1\times, 0.5\times\}$ & 22.43$\pm$0.36 & {27.77$\pm$0.18} & {32.50$\pm$0.08} & {36.06$\pm$0.24} \\
    
    \midrule
    MixTeacher-RD (Ours) & $\{1\times, 0.5\times\}$ & 23.61$\pm$0.38 & 28.45$\pm$0.16 &  {33.64$\pm$0.12}  & {36.57$\pm$0.20}  \\

    MixTeacher (Ours) & $\{1\times, 1\times, 0.5\times\}$ & \textbf{25.16$\pm$0.26} & \textbf{29.11$\pm$0.21} & \textbf{34.06$\pm$0.13} & \textbf{36.72$\pm$0.16}  \\
     \bottomrule
    \end{tabular}
    }
    
    \caption{Comparison with state-of-the-art methods on COCO benchmark. $\text{AP}_{50:95}$ on \texttt{val2017} set are reported.
    Under the {Partially Labeled} setting, results are the average of all five folds and numbers behind $\pm$ indicate the standard deviation. Under the {Additional} setting, numbers in front of the arrow indicate the supervised baseline. 
    The views of unlabeled image used in each iteration are reported as well.
    }
    \label{tab:sota_coco}
    
\end{table*}

\section{Quality of Pseudo Labels}

We further investigate the quality of pseudo labels during training. We evaluate the pseudo labels over 5,000 unlabeled images every 10k iterations for all methods. \cref{fig:quality} (a) shows the precision of pseudo labels for the proposed MixTeacher and baseline with the same score threshold of 0.9.  As the results show in~\ref{fig:quality} (a), our method obviously produces more accurate pseudo-labels, and thus achieves more accurate results in the end. \cref{fig:quality} (b) shows the precision of the pseudo labels in different range of confidence score and the pseudo labels mined by our PLM module. Compared with all the pseudo labels with thresholds in $[0.7, 0.9)$, the pseudo labels mined by our method also have higher accuracy.

\end{document}